\documentclass[conference]{IEEEtran}
\usepackage{times}
\usepackage{graphicx}
\usepackage[bookmarks=true]{hyperref}
\usepackage[
  backend=biber,
  style=numeric,
  sorting=nyt,
  natbib=true
]{biblatex}
\hypersetup{
  pdfauthor={Rohit Mohan},
  pdftitle={Perception Beyond the Visible: Incompleteness-Aware Scene Understanding for Robot Autonomy}
}

\begin{document}

\title{Perception Beyond the Visible: Incompleteness-Aware Scene Understanding for Robot Autonomy}
\author{Rohit Mohan}

\maketitle
\begingroup
\renewcommand\thefootnote{1}
\footnotetext{To acknowledge the contributions of my co-authors, I use “we” where appropriate. I use “pp” to denote percentage points. For brevity, I omit discussion of my contributions to the benchmarks and the evaluation metrics introduced in Sec.~II.}
\endgroup

\section{Introduction}

Robots do not perceive the world directly but they infer it from partial, noisy measurements~\cite{thrun2005probabilistic}. Reliable robot autonomy requires converting these measurements into a scene belief that is (i) complete enough to support safe action and (ii) calibrated enough to expose when it may be wrong~\cite{guo2017calibration,kendall2017uncertainties}. Yet much of today’s perception reliability is achieved by engineering away uncertainty: operating in bounded environments, with fixed ontologies, and with clean, well-calibrated sensors~\cite{siciliano2016springerhandbookrobotics, koopman2016challenges}. In everyday operation, three forms of incompleteness break these assumptions: incomplete evidence from occlusion and unobserved future states, incomplete knowledge in open worlds (novel objects and limited labels)~\cite{bendale2016towards}, and incomplete reliability when sensors degrade or disagree~\cite{feng2020deep}. Under these conditions, models often produce not only errors but overconfident hypotheses~\cite{guo2017calibration,ovarazquez2021confidence} that propagate into brittle downstream behavior, leading to planning failures under perceptual uncertainty~\cite{kalra2016driving,chen2015planning}.

My research program, Perception Beyond the Visible, asks: \textbf{how can robots build scene representations over semantics, geometry, and motion that remain calibrated under incompleteness and actionable for planning and interaction.} I pursue this through a tight loop of (1) tasks and benchmarks~\cite{fong2022panoptic, mohan2022amodal, sekkat2024amodalsynthdrive, li2025amodal, mohan2024panoptic, mohan2025open, mohan2026upfuse} that make incompleteness measurable, and (2) learning methods that explicitly model occlusion structure~\cite{mohan2022amodal, mohan2022perceiving, sekkat2024amodalsynthdrive, luz2024amodal} forecast future states~\cite{mohan2026forecastocc}, separate known from unknown~\cite{mohan2024panoptic, mohan2025open}, learn from limited labels~\cite{hindel2025label, hurtado2025learning, mohan2024syn}, and estimate modality reliability for selective fusion~\cite{mohan2026upfuse, mohan2024progressive}. Across 2D and 3D, these contributions shift perception from \textit{what is seen} to \textit{what is present}, widening the operational envelope in cluttered, dynamic settings.

\section{Current Research}
\noindent\textbf{Foundational Contributions.} To facilitate robust autonomy, I establish methodological foundations that make progress measurable and support rapid development. First, we developed a large-scale multimodal benchmark and instance-centric metrics for LiDAR panoptic tracking~\cite{fong2022panoptic}. With over $13\times$ more instances than SemanticKITTI, its scale and urban diversity expose failure modes in dense, dynamic scenes and underpin my subsequent work~\cite{hindel2025label, mohan2025open, mohan2026upfuse}. Second, complementing classification-centric NAS surveys~\cite{elsken2019neural}, we systematized NAS for \emph{dense prediction}~\cite{mohan2023neural}, distilling pixel-level constraints and architectural insights I leverage to design efficient networks across my core research~\cite{mohan2026forecastocc, mohan2026upfuse}. Together, these contributions enable my overarching vision.

\begin{figure}[t]
  \centering
  \begin{minipage}[t]{0.49\linewidth}
    \vspace{0pt}
    \centering
    {\footnotesize (A) Amodal Panoptic Segmentation}\\[0.15em]
    \includegraphics[width=\linewidth]{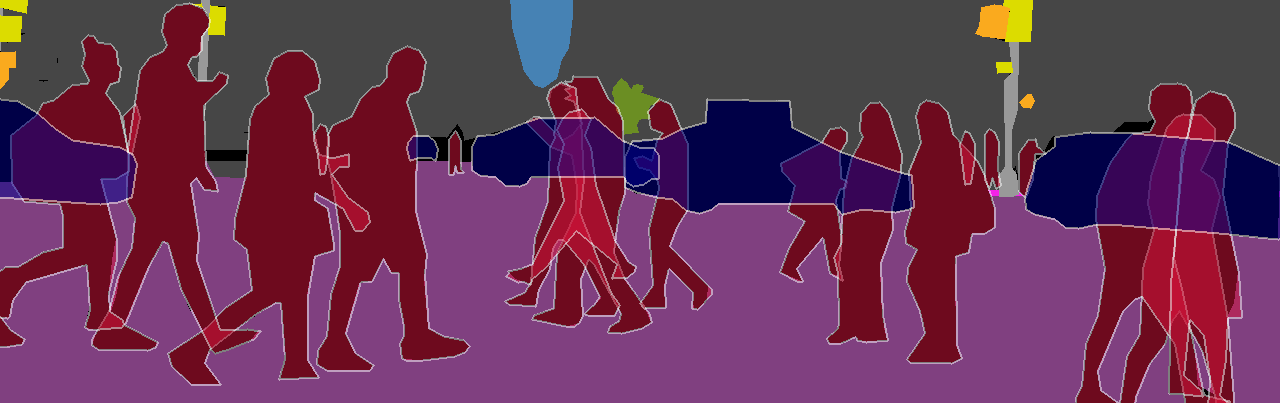}\\[0.4em]
    {\footnotesize (B) Panoptic Out-of-Distribution Segmentation}\\[0.15em]
    \includegraphics[width=\linewidth]{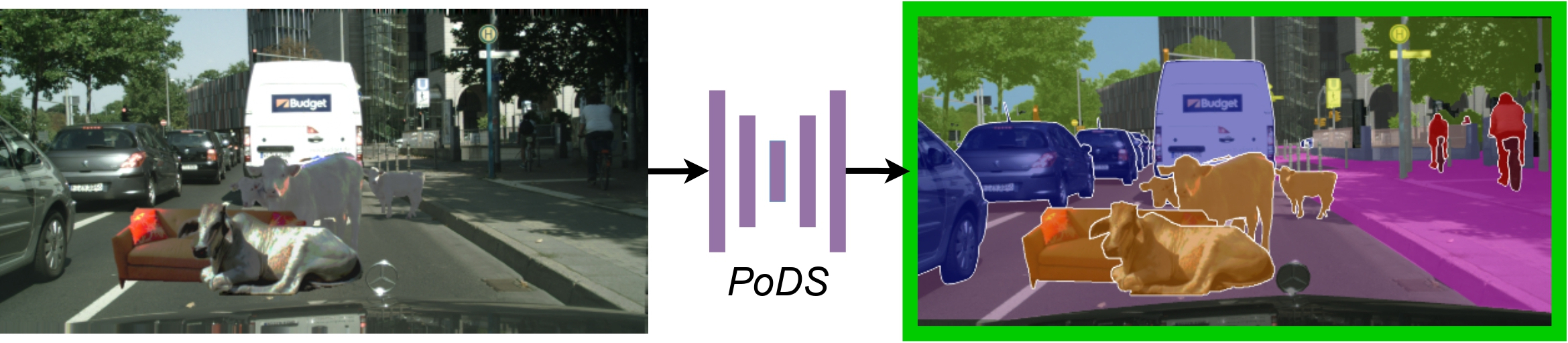}
  \end{minipage}\hfill
  \begin{minipage}[t]{0.49\linewidth}
    \vspace{0pt}
    \centering
    {\footnotesize (C) Degradation under Sensor Failure}\\[0.75em]
    \includegraphics[width=\linewidth]{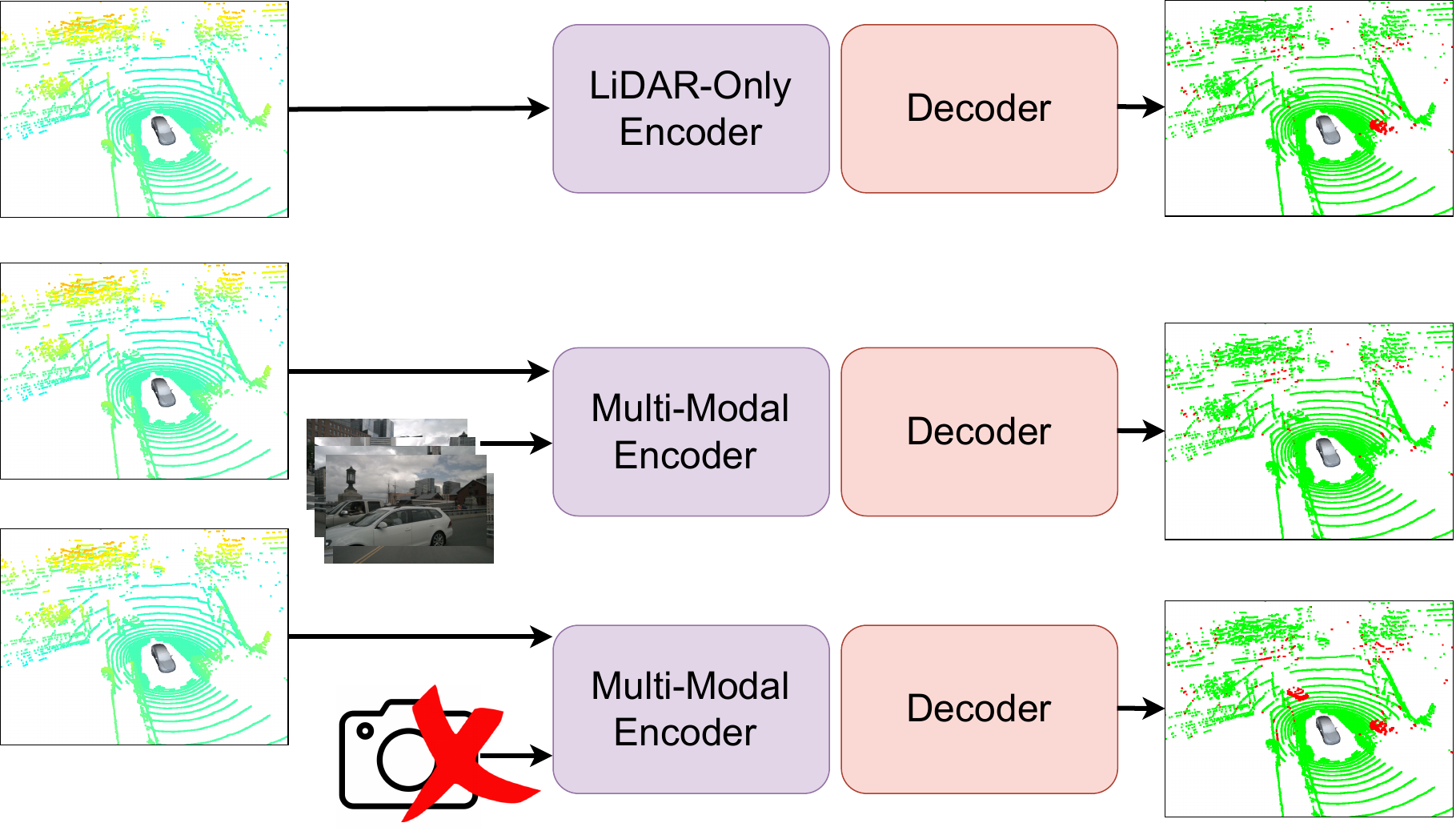}
  \end{minipage}
  \caption{\footnotesize Illustrations of incompleteness-aware perception: (A) occlusion, (B) unknowns, and (C) sensor failure.}
  \label{fig:teaser}
  \vspace{-1.2em}
\end{figure}

\subsection{Incomplete Evidence: Perceiving the Invisible}
\noindent\textbf{Amodal Perception.} Most robotic perception remains visible-only (modal)~\cite{kirillov2019panoptic, cheng2020panoptic},
treating occlusion as missing evidence rather than an inferable scene state~\cite{hoiem2007recovering,tighe2014occlusionordering}.
I formalize \emph{amodal perception} so robots can represent what is present, not just what is seen, supporting object permanence for interaction~\cite{peng2024opf}.
I introduced \emph{amodal panoptic segmentation}~(APS) and benchmark~\cite{mohan2022amodal}, unifying background semantics with instance masks that include occluded regions (cf. Fig.~\ref{fig:teaser} (A)) and extending earlier amodal instance/semantic formulations~\cite{li2016amodal,zhu2017semantic} for a holistic scene interpretation.
My initial APS approach explicitly predicts visible/occluding/occluded regions to capture occluder-occludee structure~\cite{mohan2022amodal}.
To address severe occlusion, I proposed a \emph{proposal-free} layered formulation that predicts non-overlapping masks ordered by occlusion, together with an explicit-memory refiner that completes occluded regions via experience-based shape embeddings~\cite{mohan2022perceiving}.
This improves occluded-object segmentation by $\sim$5\,pp and transfers to amodal instance segmentation with comparable gains.
Since annotating invisible regions is subjective and costly, we curated \textit{AmodalSynthDrive}~\cite{sekkat2024amodalsynthdrive}, a large synthetic dataset with amodal ground truth, and introduced \textit{amodal depth} estimation to infer occluded geometry.
We further extended amodal reasoning to dynamics with \textit{amodal optical flow}~\cite{luz2024amodal}, predicting motion through occlusions rather than failing at occlusion boundaries. These resources have since been adopted for occlusion-aware adaptation and amodal depth pretraining~\cite{oass_eccv2024,unlock_iccv2025,amodal_depth_anything_iccv2025}.
Coupling APS with amodal flow improves panoptic tracking and reduces ID switches during crossings.\looseness=-1

\noindent\textbf{Forecasting.}
Autonomy also faces incomplete evidence in time: a robot must act before the future is observed. Building on my work on reactive panoptic tracking~\cite{hurtado2025learning}, I introduced an end-to-end framework for vision-based semantic occupancy forecasting~\cite{mohan2026forecastocc}. In contrast to two-stage pipelines that first estimate an intermediate occupancy grid and then forecasts~\cite{liao2025i2,gu2024dome}, my approach forecasts future geometry and semantics directly in the multi-view feature space. This yields a $\sim$10$\times$ speedup while matching or slightly improving accuracy at a 3\,s horizon. The resulting representation is directly queryable by planners, supporting safer interaction in dynamic environments.\looseness=-1

\vspace{-0.5em}

\subsection{Incomplete Knowledge: Learning Beyond the Known}

\noindent\textbf{Open-world unknowns.}
Robots cannot rely on fixed ontologies: objects absent from training can appear at deployment. To address this holistically, I introduced panoptic out-of-distribution (OOD) segmentation~\cite{mohan2024panoptic}, augmenting panoptic parsing with an explicit unknown category (cf.~Fig.~\ref{fig:teaser} (B). In contrast to open-set panoptic settings that may expose unknown regions during training via void labels~\cite{xu2022dual,hwang2021exemplar}, my formulation evaluates unknown instances that are fully held out during training. I proposed a discrepancy-driven framework that separates unknown representations from a frozen closed-world reference, improving unknown handling
while preserving in-distribution performance (mitigating a common trade-off). I extended this to 3D via an uncertainty-guided LiDAR open-set panoptic framework~\cite{mohan2025open}. By deriving unknowns to emerge from evidential learning~\cite{sensoy2018evidential} and uncertainty-gated grouping, rather than feature-space clustering~\cite{wong2020identifying} or fixed proxies~\cite{chakravarthy2025lidar}, this approach yields state-of-the-art unknown segmentation quality. 

\noindent\textbf{Limited Supervision.}
While foundation models~\cite{caron2021emerging,oquab2023dinov2} thrive on large image corpora, comparable resources are scarce for LiDAR and hyperspectral imagery. We develop adapter-style transfer from frozen vision backbones to label-scarce modalities~\cite{hindel2025label,hurtado2026hyperspectral}. For LiDAR-only segmentation, we introduce a 2D-3D cross-attention adapter that injects Bird's-Eye-View geometry into range-view features obtained from a frozen backbone. Unlike camera-LiDAR distillation methods~\cite{liu2023segment,xu20244d}, this requires no synchronized RGB. With limited labels, it improves mIoU by +5.2/+3.0\,pp over the strongest vision-distillation baselines across benchmarks~\cite{liu2023segment,chen2023clip2scene}. For hyperspectral segmentation~\cite{hurtado2026hyperspectral}, an adapter preserves full spectra and aligns them with frozen vision tokens, yielding up to +14.6\,pp mIoU over prior works~\cite{theisen2024hs3,gutierrez2023hsi,winkens2017hyko}. To support privacy-constrained domains, I curated a synthetic multimodal dataset~\cite{mohan2024syn} with dense ground truth and showed that pairing domain-specific synthetic data
with strong vision models yields accurate, domain-aligned perception with minimal real data.\looseness=-1
\subsection{Incomplete Reliability: Robust Multi-Modal Perception}
I move beyond LiDAR-camera fusion pipelines that implicitly assume clean inputs toward models that explicitly reason about modality reliability. In 3D panoptic segmentation, I identify a recurring failure mode: fusing a corrupted modality can affect performance relative to using a single robust sensor(cf Fig.~\ref{fig:teaser} (C)). Such corruptions arise from sensor failures and calibration drift, and can induce large drops (e.g., 5-16\,pp) in PQ~\cite{kirillov2019panoptic} for state-of-the-art approaches~\cite{pan2025images, zhang2023lidar}. To address this, I propose a reliability-aware fusion mechanism~\cite{mohan2026upfuse}. By estimating aleatoric uncertainty via representational divergence, the model decides \emph{when} to trust each modality and down-weights unreliable visual features to avoid contaminating geometric reasoning. Under ideal conditions, it remains comparable to strong fusion baseline~\cite{pan2025images} while being 6$\times$ faster. More importantly, under sensor failures, sensor failures, it drops by only 1.2\,pp (vs.\ \(\sim\)5\,pp for the IAL baseline~\cite{pan2025images}) and remains 4.5-12.6\,pp ahead under rotation-induced calibration drift~\cite{pan2025images, zhang2023lidar}. Complementarily, to handle severe sensor dropout in 3D object detection, I develop a progressive fusion framework with masked cross-modal pre-training, enabling the
model to reconstruct missing features from remaining modalities~\cite{mohan2024progressive}. Together, these contributions make multi-modal perception robust by design rather than only accurate in ideal conditions.
\section{Future Research Directions}
To advance closed-loop autonomy, the next phase of my research focuses on incompleteness-aware scene belief estimation. I plan to explore two primary themes:

\noindent\textbf{Belief-centric 4D scene representations.} Instead of treating perception outputs as point estimates, I will develop methods that learn representations for jointly modeling 3D geometry, semantics, and motion alongside uncertainty (occlusion, unknowns, sensor reliability). A key direction is coupling layered amodal reasoning with semantic occupancy forecasting. As occupancy is inherently amodal, this naturally unifies \textit{object permanence} and \textit{future state} into a single scene belief. Evaluation will focus on calibration, robustness, and downstream planning in dynamic clutter.

\noindent\textbf{Closing the Loop: Active Sensing and Safe Adaptation.} Once a robot can quantify incompleteness, it must act to resolve it behaviorally and algorithmically. First, I will study active perception policies that adapt viewpoints, sensing modalities, and motion plans to maximize expected information gain, for instance, preferring geometric sensing when visual inputs degrade. This requires uncertainty estimates that are decision, relevant and stable under shift. Algorithmically, I will explore continual learning pipelines that safely incorporate newly discovered unknowns over time. By efficiently acquiring supervision for novel object hypotheses without catastrophic forgetting, the robot will expand its open-world coverage while maintaining conservative behavior when knowledge is incomplete. Evaluation will pair controlled incompleteness suites (occlusion, corruption, novel encounters) with embodied tasks where the robot explicitly trades time and energy for uncertainty reduction.

Together, these directions operationalize Perception Beyond the Visible for reliable robot autonomy.

\printbibliography

\end{document}